\documentclass{article}

\usepackage{microtype}
\usepackage{graphicx}
\usepackage{subfigure}
\usepackage{booktabs} 

\usepackage{hyperref}

\usepackage[accepted]{icml2024}

\usepackage{amsmath}
\usepackage{amssymb}
\usepackage{mathtools}
\usepackage{amsthm}

\usepackage[capitalize,noabbrev]{cleveref}

\theoremstyle{plain}

\theoremstyle{definition}

\theoremstyle{remark}

\usepackage{amsmath,amsfonts,bm}

\def\eqref#1{equation~\ref{#1}}

\def\1{\bm{1}}

\def\rvz{{\mathbf{z}}}

\def\vx{{\bm{x}}}

\def\mX{{\bm{X}}}

\DeclareMathAlphabet{\mathsfit}{\encodingdefault}{\sfdefault}{m}{sl}
\SetMathAlphabet{\mathsfit}{bold}{\encodingdefault}{\sfdefault}{bx}{n}

\def\gT{{\mathcal{T}}}

\def\sL{{\mathbb{L}}}

\def\sV{{\mathbb{V}}}

\usepackage[textsize=tiny]{todonotes}

\icmltitlerunning{Structured Generations: Using Hierarchical Clusters to guide Diffusion Models}

\begin{document}

\twocolumn[
\icmltitle{Structured Generations: Using Hierarchical Clusters to guide Diffusion Models}



\icmlsetsymbol{equal}{*}

\begin{icmlauthorlist}
\icmlauthor{Jorge da Silva Gonçalves}{yyy}
\icmlauthor{Laura Manduchi}{yyy}
\icmlauthor{Moritz Vandenhirtz}{yyy}
\icmlauthor{Julia E. Vogt}{yyy}
\end{icmlauthorlist}

\icmlaffiliation{yyy}{Department of Computer Science, ETH Zurich, Zurich, Switzerland}

\icmlcorrespondingauthor{Jorge da Silva Gonçalves}{jorge.dasilvagoncalves@inf.ethz.ch}

\icmlkeywords{Machine Learning, ICML}

\vskip 0.3in
]



\printAffiliationsAndNotice{\icmlEqualContribution} 

\begin{abstract}
This paper introduces Diffuse-TreeVAE, a deep generative model that integrates hierarchical clustering into the framework of Denoising Diffusion Probabilistic Models (DDPMs).
The proposed approach generates new images by sampling from a root embedding of a learned latent tree VAE-based structure, it then propagates through hierarchical paths, and utilizes a second-stage DDPM to refine and generate distinct, high-quality images for each data cluster. The result is a model that not only improves image clarity but also ensures that the generated samples are representative of their respective clusters, addressing the limitations of previous VAE-based methods and advancing the state of clustering-based generative modeling.
\end{abstract}

\section{Introduction}
\label{introduction}

Generative modeling and clustering represent two fundamental and distinct approaches within the field of machine learning. Generative modeling aims to approximate the underlying distribution of data, thereby enabling the generation of new samples \cite{kingma_auto-encoding_2013, goodfellow_generative_2014}. Clustering, conversely, seeks to identify meaningful and interpretable structures within data. This is achieved through the unsupervised detection of intrinsic relationships and dependencies \cite{ezugwu_comprehensive_2022}, which can enhance data visualization and interpretation.
TreeVAE \cite{manduchi_tree_2023} was recently proposed to combine these two research directions by integrating hierarchical dependencies into a deep latent variable model. TreeVAE models the distribution of data by learning the optimal tree structure of latent variables. The resulting latent embeddings are automatically organized into a hierarchical structure that mimics the hierarchical clustering process.  
As a result, it can generate new data via conditional sampling and perform hierarchical clustering. However, its generative performance falls short of state-of-the-art deep generative methods, and it exhibits common issues associated with VAEs, such as generating blurry images \cite{bredell_explicitly_2023}. In contrast, diffusion models \cite{sohl-dickstein_deep_2015, ho_denoising_2020} have recently gained significant attention for their image-generation capabilities.

\begin{figure*}[ht]
\begin{center}
\centerline{\includegraphics[width=0.9\textwidth]{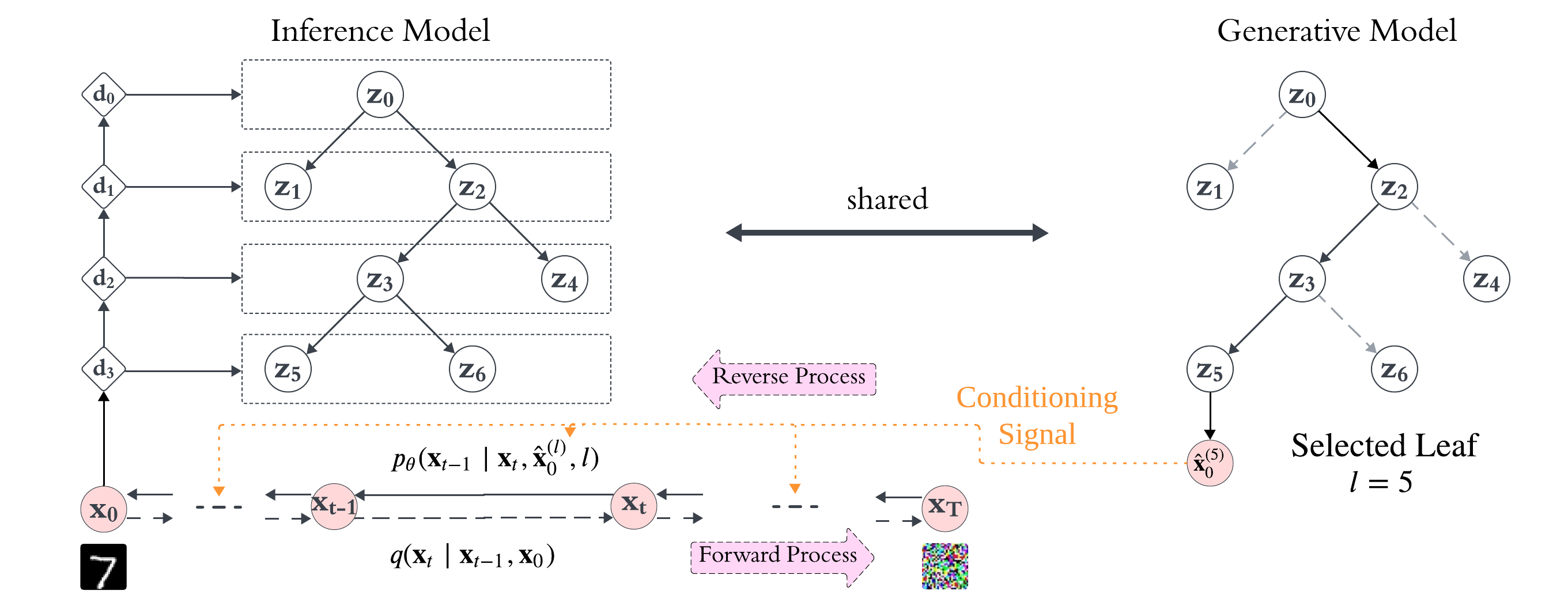}}
\caption[Schematic overview of the Diffuse-TreeVAE model]{Schematic overview of the Diffuse-TreeVAE model: The reverse model of the DDPM (bottom) is conditioned on both the reconstruction and the index of the selected leaf $l$ obtained from the associated, pre-trained TreeVAE. The denoising function of the DDPM learns to refine the TreeVAE-based reconstructions.}
\label{diff-treevae}
\end{center}
\vskip -0.2in
\end{figure*}

Our work aims to bridge this gap by (a) improving the architectural design of TreeVAE, and (b) integrating a second-stage Denoising Diffusion Probabilistic Model (DDPM) that is conditioned on the cluster-specific representations learned by TreeVAE. Our proposed approach, Diffuse-TreeVAE, generates high-quality, distinct, and representative cluster-specific images; i.e. images for each leaf of the learned tree. The proposed generation process (depicted in \cref{diff-treevae}) goes as follows: Diffuse-TreeVAE samples the root embedding of its tree, then propagates the generations through the leaves, and, finally, it produces high-quality leaf-specific images by guiding the reverse process of the DDPM on both the leaf reconstructions and the corresponding path in the tree. 
The resulting leaf-specific images share common general properties (which are sampled at the root) and differ by cluster-specific features.

\textbf{Our main contributions} are as follows: We provide (i) a holistic approach to clustering-based generative modeling, and (ii) a novel method for controlling image synthesis in diffusion models. We show that our approach (a) overcomes previous generative limitations of VAE-based clustering methods, and (b) produces newly generated samples that are more representative of the respective clusters in the data and closer to the true image distribution.

\section{Diffuse-TreeVAE}
\label{model}

We propose Diffuse-TreeVAE\footnote{The code is publicly available at \url{https://github.com/JoGo175/diffuse-treevae}}, a two-stage generative framework that is composed of a VAE-based generative hierarchical clustering model (TreeVAE), followed by a cluster-conditional denoising diffusion probabilistic model (DDPM).  This novel combination of VAEs and diffusion models extends the generator-refiner framework introduced by DiffuseVAE \cite{pandey_diffusevae_2022} to hierarchical clustering tasks. Here, TreeVAE \cite{manduchi_tree_2023} serves as the generator, while a DDPM \cite{ho_denoising_2020}, conditioned on the TreeVAE leaves, refines the generated images. \cref{diff-treevae} illustrates the workflow of Diffuse-TreeVAE.

The first part of Diffuse-TreeVAE involves an adapted version of the TreeVAE model \cite{manduchi_tree_2023}. TreeVAE is a generative model that intrinsically learns to hierarchically separate data into clusters via a latent tree. During training, the model grows a binary tree structure $\gT$. The set $\sV$ represents the nodes of the tree. Each node corresponds to a stochastic latent variable, denoted as $\rvz_0, \dots, \rvz_V$. The parameters of these latent variables are determined by their parent nodes through neural networks called transformations. The set of leaves $\sL$, where $\sL \subset \sV$, represents the clusters present in the data. Starting from the root node, $\rvz_0$, a given sample traverses the tree to a leaf node, $\rvz_l$, in a probabilistic manner. The probabilities for whether to go to the left or right child at each internal node are determined by neural networks termed routers. Thus, the latent tree encodes sample-specific probability distributions of paths. Each leaf embedding, $\rvz_l$ for $l \in \sL$, represents the learned data representations, and leaf-specific decoders use these embeddings to reconstruct or generate new images, i.e. given a dataset $\mX$, TreeVAE reconstructs $\hat{\mX} = \{\hat{\mX}^{(l)} \mid l \in \sL\}$.

In this work, we improve the architectural design of the TreeVAE model. In the original TreeVAE, an initial encoder projects the images to flattened representations at the start of the bottom-up process, with the remaining components of the model relying on MLP layers. We adapted our TreeVAE method to use convolutional layers throughout the model structure instead of MLP layers. Thus, our adaptation avoids flattening the representations and instead utilizes lower-dimensional representations with multiple channels throughout the model. Additionally, we incorporated residual connections to enhance the training and performance of the model. These modifications aim to preserve spatial information and enable more efficient learning, contributing to the overall effectiveness of our model. However, it is important to note that this model suffers from the typical VAE issue of producing blurry image generations \cite{bredell_explicitly_2023}. Despite this limitation, the reconstructed images and learned clustering still provide meaningful representations of the data, which are utilized in the second stage of our proposed Diffuse-TreeVAE framework.

\begin{figure*}[t]
\begin{center}
\centerline{\includegraphics[width=0.9\textwidth]{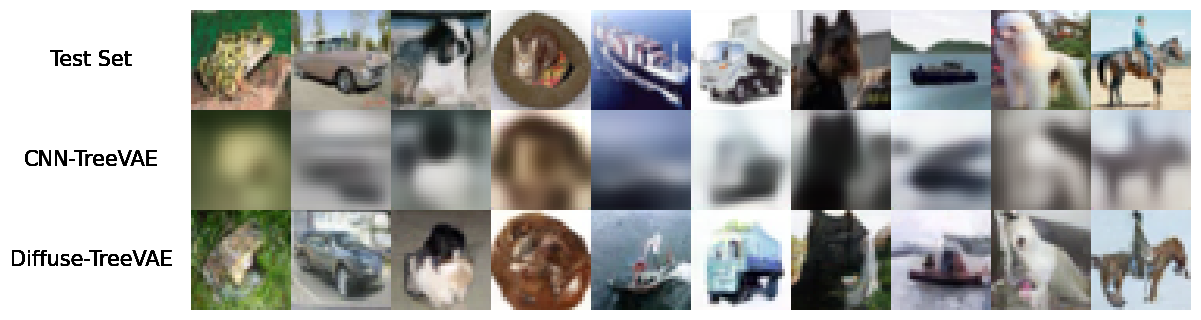}}
\caption{(Top) Samples from the CIFAR-10 test set. (Middle) Reconstructions from the CNN-TreeVAE model. (Bottom) Refined reconstructions from the Diffuse-TreeVAE model, conditioned on the CNN-TreeVAE reconstructions and the corresponding leaves.}
\label{recons_comp}
\end{center}
\vskip -0.2in
\end{figure*}

Our model leverages the cluster assignments and image generation capabilities of TreeVAE to guide a second-stage diffusion model, specifically a DDPM \cite{ho_denoising_2020}. We adopt and adapt the generator-refiner framework \cite{pandey_diffusevae_2022}, where the VAE generates the initial, typically blurred images and the conditioned DDPM refines these reconstructions to produce sharper, higher-quality images. Instead of employing a conventional VAE, our model integrates TreeVAE. During training and inference, the selected leaf is randomly sampled based on the leaf probabilities learned by TreeVAE. The selected leaf reconstruction, along with the leaf index as the cluster signal, conditions the DDPM reverse process, as depicted in \cref{diff-treevae}. 
Formally, given the input data $\mX$, here denoted as $\mX_0$, we define a sequence of $T$ noisy representations of the input $\vx_{0}$, yielding $\vx_{1:T}$. The forward process, $q\left(\vx_{1: T} \mid \vx_{0}\right)$, that gradually destroys the structure of each data sample follows the standard DDPM process \cite{ho_denoising_2020}. The reverse process, on the other hand, is conditioned on the TreeVAE reconstructions $\hat{\vx}_0 = \{\hat{\vx}_{0}^{(l)} \mid l \in \sL\}$ and on the leaf assignments:
\begin{align}
    \begin{split} 
    &l \sim p(l | \vx_0), \\
    &p_{\psi} \!\! \left(\vx_{0: T} \!\mid \!\hat{\vx}_{0}^{(l)}, l \right)=p\left(\vx_{T}\right) \prod_{t=1}^T p_\psi \!\!\left(\vx_{t-1} \!\mid \!\vx_{t}, \hat{\vx}_{0}^{(l)}, l \right)\!, \\ 
\end{split}
\end{align}
where $p(l | \vx_0)$ is the probability that the sample $\vx_0$ is assigned to leaf $l$.
This method ensures that leaves with smaller assignment probabilities are considered, encouraging the DDPM to perform effectively across all leaves. Consequently, our approach addresses the distinct clusters inherent in TreeVAE, allowing the model to adapt specifically to different clusters and encouraging cluster-specific refinements in the images. This guidance in the image generation process assists the denoising model in learning cluster-specific image reconstructions. On the other hand, the forward noising process remains unconditional. Diffuse-TreeVAE directly utilizes the reconstructions instead of the latent embeddings for conditioning, as there exists a deterministic relationship between leaf embeddings and leaf reconstruction, provided by the leaf-specific decoder.

By using the generator-refiner framework, Diffuse-TreeVAE maintains the same clustering performance as the underlying TreeVAE. The DDPM refines the generated output samples without influencing the cluster assignments in the TreeVAE model. This is achieved through a two-stage training strategy, where the conditional DDPM is trained using a pre-trained CNN-TreeVAE model. Thus, Diffuse-TreeVAE combines the effective clustering of TreeVAE with the superior image generation capabilities of diffusion models.

\section{Results}
\label{results}

We evaluate the generative performance of our model on the MNIST \cite{lecun_gradient-based_1998}, FashionMNIST \cite{xiao_fashion-mnist_2017}, and CIFAR-10 \cite{krizhevsky_learning_2009} datasets. Our analysis compares three models: the original MLP-based TreeVAE, referenced as MLP-TreeVAE \cite{manduchi_tree_2023}; our CNN-based adaptation, referred to as CNN-TreeVAE; and our novel proposal, the Diffuse-TreeVAE model. The latter model is conditioned on reconstructions and clusters derived from the CNN-TreeVAE, enhancing its capacity for generative performance. Reconstruction performance is assessed through the FID score \cite{heusel_gans_2017}, calculated for the reconstructed images sourced from the 10,000 samples within the test set. Similarly, generation performance is evaluated using the FID score, this time computed for 10,000 newly generated images. 
\begin{table}[ht]
\centering
\caption[Test set generative performances]{Test set generative performances of different TreeVAE models. Lower FID scores indicate better performance. Means and standard deviations are computed across 10 runs with different seeds. The best result for each dataset is marked in \textbf{bold}.}
\label{tab:rec_gen_performance}

    \begin{tabular*}{\linewidth}{@{\extracolsep{\fill}}llrr}
    \hline 
    \textbf{Dataset} & \textbf{Method} & \textbf{FID (rec)} & \textbf{FID (gen)} \\
    \hline MNIST    
    & MLP-TreeVAE           & $25.8 \,\text{{\footnotesize $\pm\, 0.4$}}$           & $25.3 \,\text{{\footnotesize $\pm\, 1.0$}}$  \\
    & CNN-TreeVAE         & $24.9 \,\text{{\footnotesize $\pm\, 1.1$}}$           & $22.8 \,\text{{\footnotesize $\pm\, 1.4$}}$  \\
    & Diffuse-TreeVAE       & $\textbf{1.5} \,\text{{\footnotesize $\pm\, 0.1$}}$   & $\textbf{16.2} \,\text{{\footnotesize $\pm\, 5.7$}}$ \\
                    
    \hline Fashion     
    & MLP-TreeVAE           & $44.7 \,\text{{\footnotesize $\pm\, 0.6$}}$           & $46.8 \,\text{{\footnotesize $\pm\, 0.9$}}$  \\
    & CNN-TreeVAE         & $36.5 \,\text{{\footnotesize $\pm\, 0.6$}}$           & $39.0 \,\text{{\footnotesize $\pm\, 0.8$}}$  \\
    & Diffuse-TreeVAE       & $\textbf{4.1} \,\text{{\footnotesize $\pm\, 0.6$}}$   & $\textbf{4.2} \,\text{{\footnotesize $\pm\, 0.5$}}$  \\
    \hline CIFAR10  
    & MLP-TreeVAE           & $225.5 \,\text{{\footnotesize $\pm\, 3.3$}}$          & $237.0 \,\text{{\footnotesize $\pm\, 4.0$}}$  \\
    & CNN-TreeVAE         & $190.5 \,\text{{\footnotesize $\pm\, 2.0$}}$          & $200.9 \,\text{{\footnotesize $\pm\, 2.5$}}$  \\
    & Diffuse-TreeVAE       & $\textbf{15.4} \,\text{{\footnotesize $\pm\, 0.3$}}$  & $\textbf{22.3} \,\text{{\footnotesize $\pm\, 0.3$}}$  \\
    \hline
    \end{tabular*}
\end{table}

\begin{figure*}[ht!]
\begin{center}
\centerline{\includegraphics[width=0.9\textwidth]{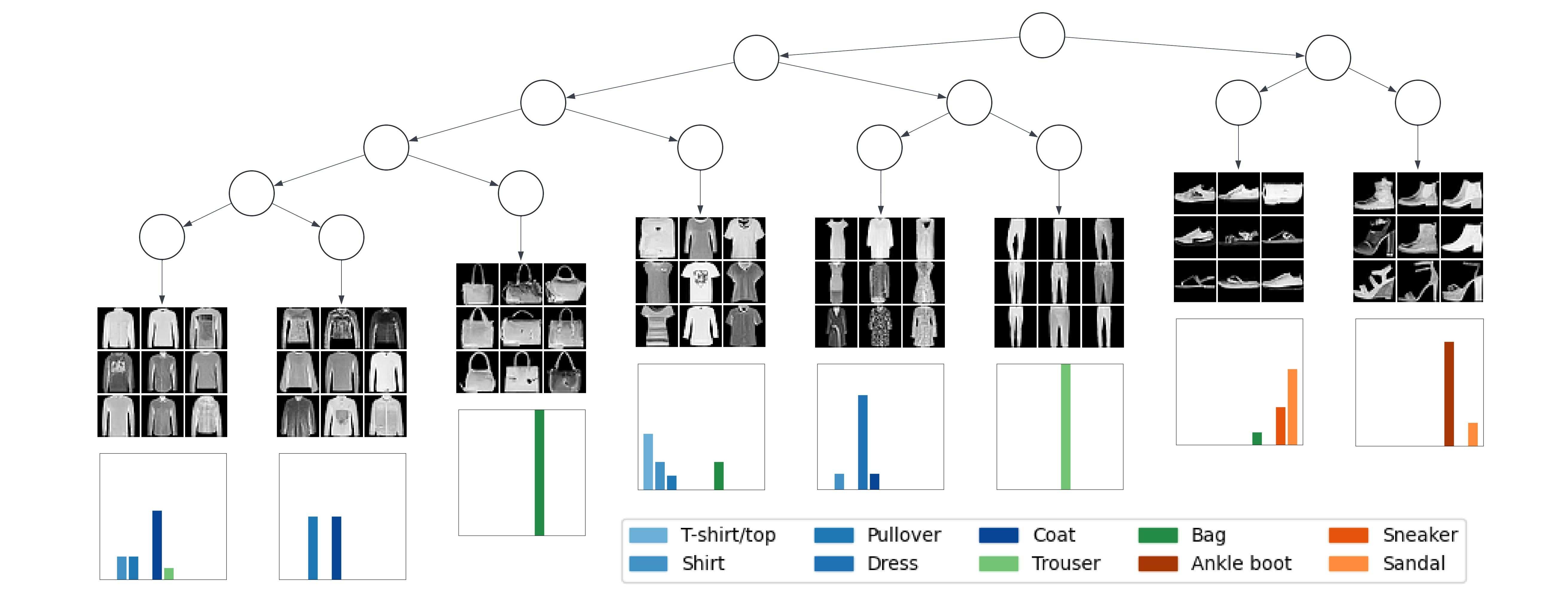}}
\caption{Diffuse-TreeVAE model trained on FashionMNIST. For each cluster, random newly generated images are displayed. Below each set of images, a normalized histogram (ranging from 0 to 1) shows the distribution of predicted classes from an independent, pre-trained classifier on FashionMNIST for all newly generated images in each leaf with a significant probability of reaching that leaf.}
\label{fmnist_generations_hist}
\end{center}
\vskip -0.3in
\end{figure*}
\begin{figure*}[ht!]
\begin{center}
\centerline{\includegraphics[width=0.9\textwidth]{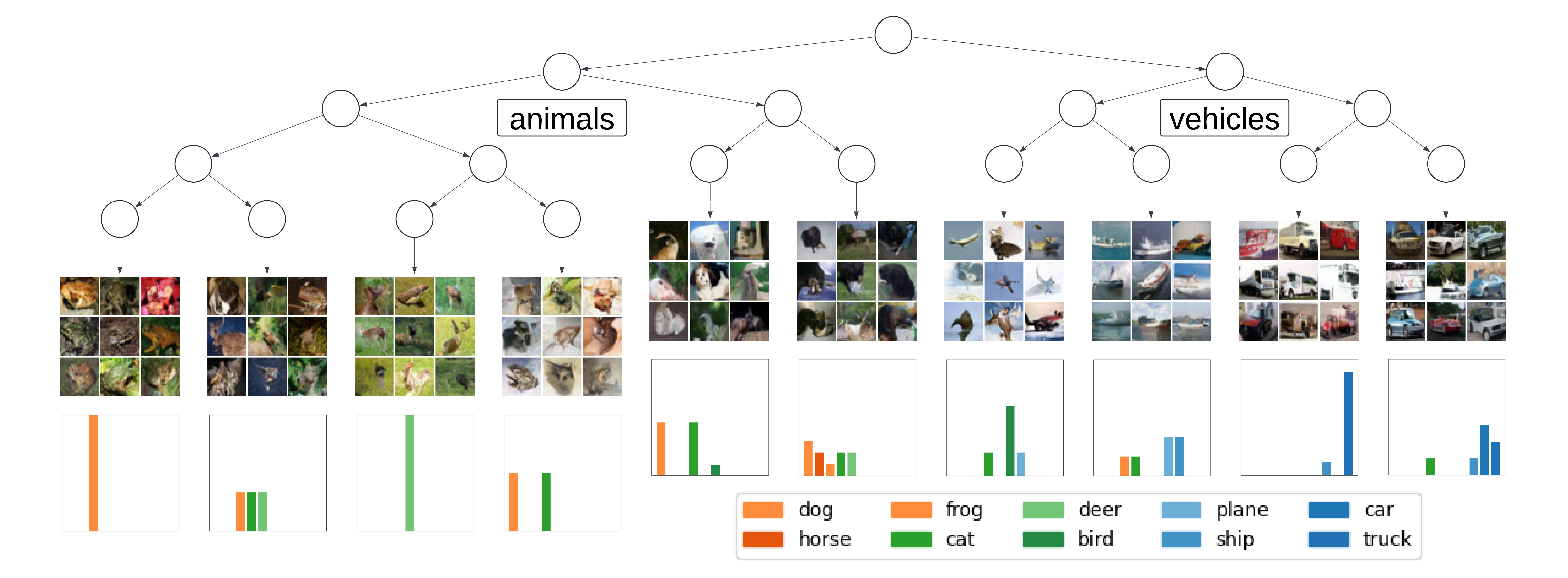}}
\caption{Diffuse-TreeVAE model trained on CIFAR-10. For each cluster, random newly generated images are displayed. Below each set of images, a normalized histogram (ranging from 0 to 1) shows the distribution of predicted classes from an independent, pre-trained classifier on CIFAR-10 for all newly generated images in each leaf with a significant probability of reaching that leaf.}
\label{cifar_generations_hist}
\end{center}
\vskip -0.3in
\end{figure*}
Table \ref{tab:rec_gen_performance} illustrates the generative performance across the various datasets. In every case, the CNN-TreeVAE demonstrates improvements compared to the original model. However, despite being lower, its FID scores remain at a similar level. Hence, the CNN-TreeVAE model continues to generate visibly blurry images. On the other hand, the Diffuse-TreeVAE significantly enhances the generative capabilities of the model, yielding much lower FID scores, often by an order of magnitude. This improvement is evident in the quality of the generated images, as depicted in \cref{recons_comp}. Here, we visually compare the reconstructions generated by the Diffuse-TreeVAE model with those produced by the underlying CNN-TreeVAE model, which was used to condition the Diffuse-TreeVAE along with the cluster signal. Specifically, it can be observed that the reconstructions generated by the Diffuse-TreeVAE are notably sharper and thus exhibit closer adherence to the true distribution of the test data. The model demonstrates enhanced capability in reproducing fine details within the images while preserving the overall color and structure. However, it is important to note that these improvements may introduce some inconsistencies, resulting in reconstructions that appear more realistic but deviate slightly from the original image being reconstructed.

To assess the quality of the newly generated images, we train a classifier on the original dataset using the training data and then utilize it to classify the newly generated images from our Diffuse-TreeVAE. Specifically, we classify the newly generated images for each cluster separately. Ideally, the majority of generated images from a cluster are classified into one or very few classes from the original dataset. The more generations from a cluster that are classified into one class only, the ``purer" or ``less ambiguous" we consider the generations to be. For this classification task, we utilize a ResNet-50 model \cite{he_deep_2016} trained on each dataset.

\begin{figure*}[ht!]
\begin{center}
\centerline{\includegraphics[width=\textwidth]{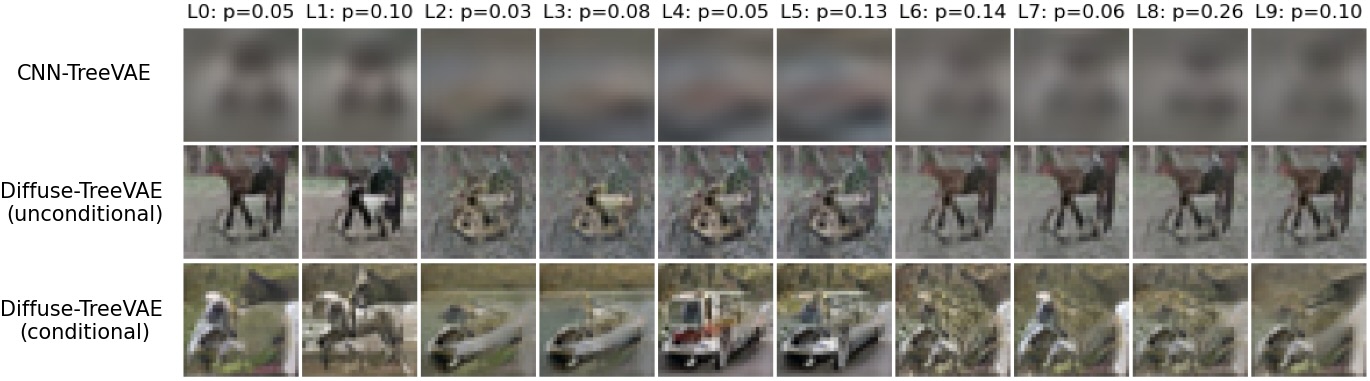}}
\caption{Image generations from each leaf of (top) a CNN-TreeVAE, (middle) a cluster-unconditional Diffuse-TreeVAE, and (bottom) a cluster-conditional Diffuse-TreeVAE, all trained on CIFAR-10. Each row displays the generated images from all leaves of the specified model, starting with the same sample from the root. The corresponding leaf probabilities are shown at the top of the image and are by design the same for all models.}
\label{cifar_generations_generations}
\end{center}
\vskip -0.3in
\end{figure*}

In \cref{fmnist_generations_hist}, we present randomly generated images from a Diffuse-TreeVAE model trained on FashionMNIST. Notably, the model in this instance has identified only seven clusters instead of the expected ten. These clusters tend to group various clothing items together, such as ``Shirt", ``T-Shirt", and ``Pullover". Below the generated images, normalized histograms depict the distribution of the predicted classes by the classifier on the newly generated images. For instance, clusters representing trousers and bags appear to accurately and distinctly capture their respective classes, as all their generated images are classified into one group only. Conversely, certain clusters manifest a mixture of classes, indicating that they are grouped together. This observation is further supported by the histograms. Similar results can be observed for the CIFAR-10 or MNIST data, as shown in \cref{cifar_generations_hist} and \cref{mnist_generations_hist} respectively.

To assess whether the additional conditioning on the selected leaf index helps create more cluster-specific representations, we perform an ablation study. This study compares the generations of two Diffuse-TreeVAE models which only differ in one aspect: one conditioned only on the reconstructions and the other conditioned on both the reconstructions and the leaf index. For this ablation, we use the previously defined independent classifier to create histograms for each leaf to evaluate how cluster-specific the newly generated images are. As previously mentioned, ideally, the majority of generated images from a cluster should be classified into one or very few classes from the original dataset. To quantify this, we compute the average entropy for all leaf-specific histograms. Lower entropy indicates less variation in the histograms, and thus more leaf-specific generations. Table \ref{tab:cond_vs_uncond} presents the results for the unconditional and conditional Diffuse-TreeVAEs across all datasets. The conditional model consistently shows lower mean entropy, indicating that additional conditioning on the leaf indices indeed helps guide the model to generate more distinct and representative images for each leaf. \cref{cifar_generations_generations} visually presents the leaf generations for one sample of these models alongside the underlying CNN-TreeVAE generations, which were used to condition both models. Further examples can be found in \ref{further_gen_examples}. It can be observed that both the unconditional and conditional models exhibit a significant improvement in image quality. However, the images in the cluster-conditional model are more diverse, demonstrating greater adaptability for each cluster. This is evident as the images clearly show indications of multiple true CIFAR-10 classes, with recognizable features such as horses, ships, or cars. Notably, across all models, the leaf-specific images share common properties sampled at the root while varying in cluster-specific features from leaf to leaf within each model.
\begin{table}[ht]
\vskip -0.075in
\centering
\caption{Cluster-specificity of Diffuse-TreeVAE generations for cluster-unconditional and cluster-conditional reverse models, measured by mean entropy. Lower entropy indicates more cluster-specific generations. Mean entropy is computed across all leaf-specific histograms of the predicted classes for newly generated images. The best result for each dataset is marked in \textbf{bold}.}
\label{tab:cond_vs_uncond}

    \begin{tabular*}{\linewidth}{@{\extracolsep{\fill}}llc}
    \hline 
    \textbf{Dataset} & \textbf{Method} & \textbf{Mean Entropy}\\
    \hline MNIST    
    & unconditional       & $1.24$\\
    & conditional         & $\textbf{0.13}$\\
    \hline Fashion     
    & unconditional       & $\textbf{0.66}$\\
    & conditional         & $\textbf{0.66}$\\
    \hline CIFAR10  
    & unconditional       & $1.12$\\
    & conditional         & $\textbf{0.82}$\\
    \hline
    \end{tabular*}
\vskip -0.1in
\end{table}

\section{Conclusion}
\label{conclusion}
In this work, we present Diffuse-TreeVAE, a novel approach to integrate hierarchical clustering into diffusion models. 
By enhancing TreeVAE with a Denoising Diffusion Probabilistic Model conditioned on the cluster-specific representations, we have developed a model capable of generating distinct, high-quality images that faithfully represent their respective data clusters. This approach not only improves the visual fidelity of generated images but also ensures that these representations are true to the underlying data distribution. Diffuse-TreeVAE offers a robust framework that bridges the gap between clustering precision and generative performance, thereby expanding the potential applications of generative models in areas requiring detailed and accurate visual data interpretation.

\section*{Acknowledgments and Disclosure of Funding}
Laura Manduchi is supported by the SDSC PhD Fellowship \#1-001568-037. Moritz Vandenhirtz is supported by the Swiss State Secretariat for Education, Research and Innovation (SERI) under contract number MB22.00047.

\clearpage

\nocite{langley00}

\bibliography{example_paper}
\bibliographystyle{icml2024}

\newpage
\appendix
\onecolumn

\section{Appendix}

\subsection{Generations on MNIST and CIFAR-10} 

\begin{figure*}[ht!]
\vskip 0.2in
\begin{center}
\centerline{\includegraphics[width=0.9\textwidth]{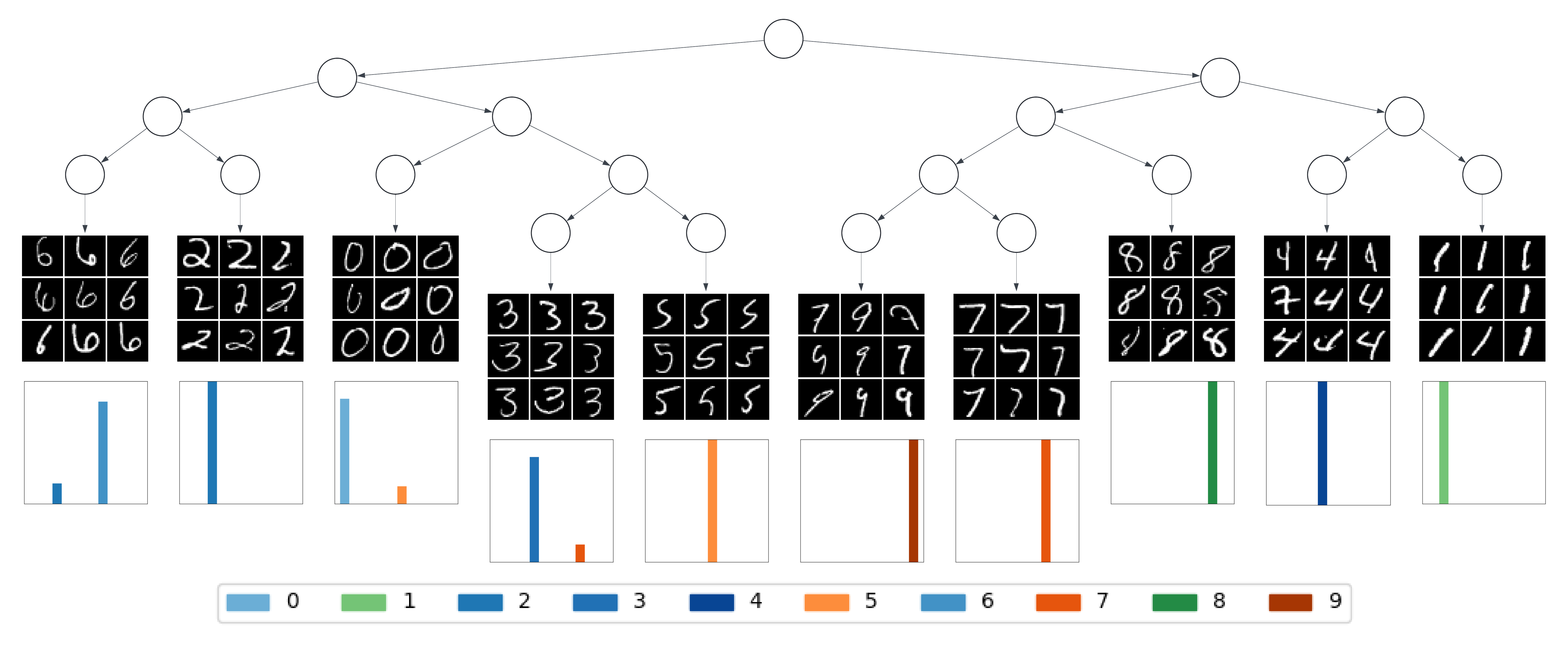}}
\vspace{-10pt} 
\caption{Diffuse-TreeVAE model trained on MNIST. For each cluster, random newly generated images are displayed. Below each set of images, a normalized histogram (ranging from 0 to 1) shows the distribution of predicted classes from an independent, pre-trained classifier on MNIST for all newly generated images in each leaf with a significant probability of reaching that leaf.}
\label{mnist_generations_hist}
\end{center}
\end{figure*}

\cref{mnist_generations_hist} presents an additional plot similar to those in \cref{fmnist_generations_hist} and \cref{cifar_generations_hist} from the main text. This plot illustrates the generated images of the Diffuse-TreeVAE model when trained on the MNIST dataset. In each of these plots, we display randomly generated images for each cluster. Below each set of leaf-specific images, we provide a normalized histogram showing the distribution of predicted classes by an independent ResNet-50 classifier that has been pre-trained on the training data of the respective dataset. This visualization helps in understanding how well the model can generate distinct and meaningful clusters in the context of different datasets.

\clearpage

\subsection{Additional Generation Examples for Conditional vs. Unconditional Diffuse-TreeVAE} \label{further_gen_examples}

\begin{figure*}[ht!]
\begin{center}
\centerline{\includegraphics[width=\textwidth, trim=120 0 90 10, clip]{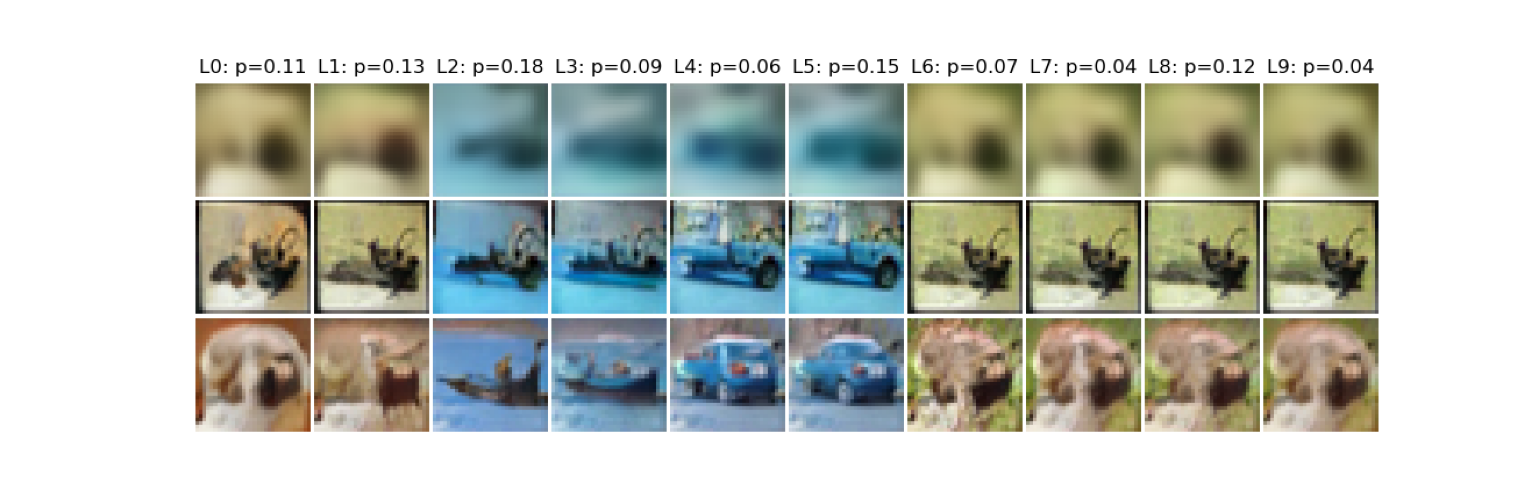}}
\centerline{\includegraphics[width=\textwidth, trim=120 0 90 40, clip]{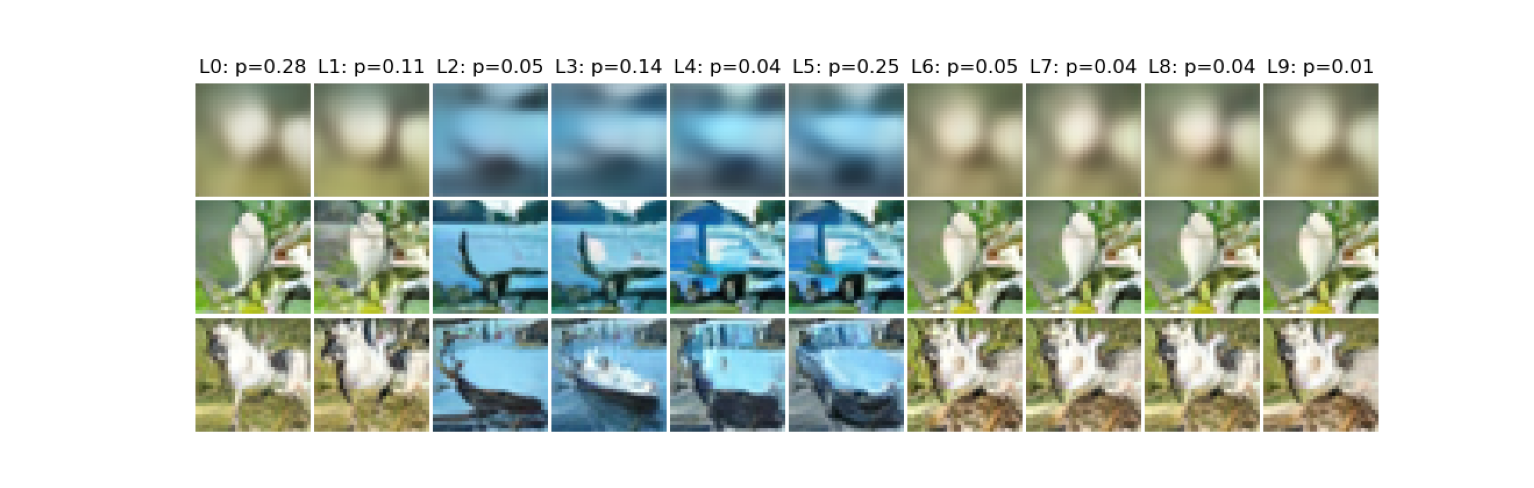}}
\centerline{\includegraphics[width=\textwidth, trim=120 40 90 40, clip]{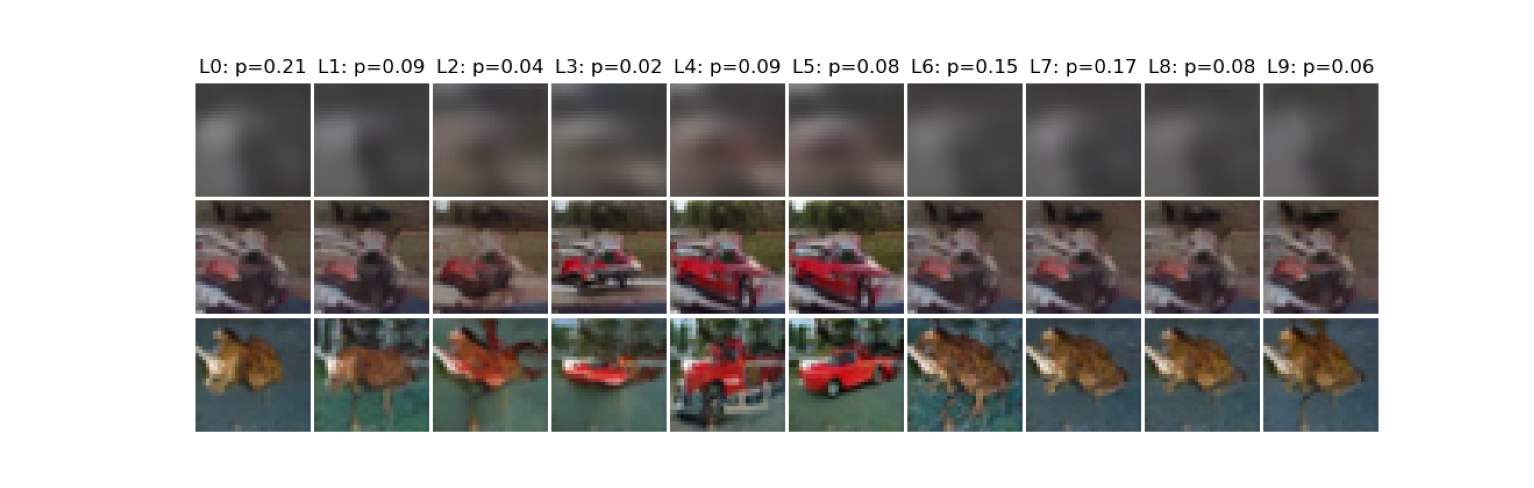}}
\vspace{-10pt} 
\caption{For each example, we show image generations from each leaf of (top) a CNN-TreeVAE, (middle) a cluster-unconditional Diffuse-TreeVAE, and (bottom) a cluster-conditional Diffuse-TreeVAE, all trained on CIFAR-10. Each row displays the generated images from all leaves of the specified model, starting with the same sample from the root. The corresponding leaf probabilities are shown at the top of the image and are by design the same for all models.}
\label{cifar_generations_generations1}
\end{center}
\vskip -0.4in
\end{figure*}

\cref{cifar_generations_generations1} presents three additional examples similar to \cref{cifar_generations_generations}, comparing the leaf-specific generations of the conditional and unconditional Diffuse-TreeVAE models, alongside the underlying CNN-TreeVAE generations.

\end{document}